\documentclass[11pt,a4paper]{article}
\usepackage[hyperref]{acl2020}
\usepackage{times}
\usepackage{capt-of}
\usepackage{enumitem}
\usepackage{latexsym}

\usepackage{amsfonts}
\usepackage[utf8]{inputenc}
\usepackage{url}
\usepackage{latexsym}
\usepackage{covington}
\usepackage{booktabs}
\usepackage{relsize}
\usepackage{todonotes}
\usepackage{xspace}
\usepackage{amsmath}
\usepackage{hyperref}
\usepackage{multirow}
\usepackage[all]{nowidow}
\usepackage{breakurl}
\usepackage{color}

\newcommand{\nerlabel}[1]{{\small \textsf{#1}}}
\newcommand{\nerlabelto}[1]{{\scriptsize \textsf{#1}}}
\newcommand{\labelfunsmall}[1]{{\scriptsize \textsf{#1}}}
\usepackage{arydshln}
\usepackage{amsmath}

\DeclareMathOperator*{\argmin}{arg\,min}

\usepackage{microtype}

\aclfinalcopy 
\def\aclpaperid{2772}

\title{Named Entity Recognition without Labelled Data:\\A Weak Supervision Approach}

\author{Pierre Lison \and Aliaksandr Hubin\\
  Norwegian Computing Center \\
  Oslo, Norway \\
  \texttt{\{plison,ahu\}@nr.no} \\\And
  Jeremy Barnes \and Samia Touileb \\
  Language Technology Group \\
  University of Oslo, Norway \\
  \texttt{\{jeremycb,samiat\}@ifi.uio.no}}
\date{}

\begin{document}
\maketitle
\begin{abstract}
  Named Entity Recognition (NER) performance often degrades rapidly when applied to target domains that differ from the texts observed during training. When in-domain labelled data is available, transfer learning techniques can be used to adapt existing NER models to the target domain. But what should one do when there is no hand-labelled data for the target domain? This paper presents a simple but powerful approach to learn NER models in the absence of labelled data through \textit{weak supervision}. The approach relies on a broad spectrum of labelling functions to automatically annotate texts from the target domain. These annotations are then merged together using a hidden Markov model which captures the varying accuracies and confusions of the labelling functions. A sequence labelling model can finally be trained on the basis of this unified annotation. We evaluate the approach on two English datasets (CoNLL 2003 and news articles from Reuters and Bloomberg) and demonstrate an improvement of about 7 percentage points in entity-level $F_1$ scores compared to an out-of-domain neural NER model.
  
\end{abstract}

\section{Introduction}
\label{sec:intro}

Named Entity Recognition (NER) constitutes a core component in many NLP pipelines and is employed in a broad range of applications such as information extraction \cite{Raiman2018DeepTypeME}, question answering \cite{molla-etal-2006-named}, document de-identification \cite{Stubbs:2015:ASD:2869975.2870353}, machine translation \cite{ugawa-etal-2018-neural} and even conversational models \cite{ghazvininejad2018a}. Given a document, the goal of NER is to identify and classify spans referring to an entity belonging to pre-specified categories such as persons, organisations or geographical locations. 

NER models often rely on convolutional or recurrent neural architectures, sometimes completed by a CRF layer \cite{chiu-nichols-2016-named,lample-etal-2016-neural,yadav-bethard-2018-survey}. More recently, deep contextualised representations relying on bidirectional LSTMS \cite{peters-etal-2018-deep},  transformers \cite{devlin-etal-2019-bert,yan2019tener} or contextual string embeddings \cite{akbik-etal-2019-pooled} have also been shown to achieve state-of-the-art performance on NER tasks.  

These neural architectures require large corpora annotated with named entities, such as Ontonotes \cite{ontonotes2011} or ConLL 2003 \cite{tjong-kim-sang-de-meulder-2003-introduction}. When only modest amounts of training data are available, transfer learning approaches can transfer the knowledge acquired from related tasks into the target domain, using techniques such as simple transfer \cite{rodriguez-etal-2018-transfer}, discriminative fine-tuning \cite{howard-ruder-2018-universal},
adversarial transfer \cite{zhou-etal-2019-dual} or layer-wise domain adaptation approaches \cite{yang2017transfer,lin-lu-2018-neural}.

However, in many practical settings, we wish to apply NER to domains where we have no labelled data, making such transfer learning methods difficult to apply. This paper presents an alternative approach using \textit{weak supervision} to bootstrap named entity recognition models without requiring any labelled data from the target domain.  The approach relies on labelling functions that automatically annotate documents with named-entity labels. A hidden Markov model (HMM) is then trained to unify the noisy labelling functions into a single (probabilistic) annotation, taking into account the accuracy and confusions of each labelling function. Finally, a sequence labelling model is trained using a cross-entropy loss on this unified annotation.

As in other weak supervision frameworks, the labelling functions allow us to inject \textit{expert knowledge} into the sequence labelling model, which is often critical when data is scarce or non-existent \cite{hu-etal-2016-harnessing,wang-poon-2018-deep}.  New labelling functions can be easily inserted  to leverage the knowledge sources at our disposal for a given textual domain. Furthermore, labelling functions can often be ported across domains, which is not the case for manual annotations that must be reiterated for every target domain. 

The contributions of this paper are as follows:
\begin{enumerate}[itemsep=1mm]
\item A broad collection of labelling functions for NER, including neural models trained on various textual domains, gazetteers, heuristic functions, and document-level constraints.
\item A novel weak supervision model suited for sequence labelling tasks and able to include probabilistic labelling predictions.
\item An open-source implementation of these labelling functions and aggregation model that can scale to large datasets \footnote{\url{https://github.com/NorskRegnesentral/weak-supervision-for-NER}.}.

\end{enumerate}

\section{Related Work}
\label{sec:relatedwork}

\paragraph{Unsupervised domain adaptation:}

Unsupervised domain adaptation attempts to adapt knowledge from a source domain to predict new instances in a target domain  which often has substantially different characteristics. Earlier approaches often try to adapt the feature space using \textit{pivots} \cite{blitzer-etal-2006-domain,blitzer-etal-2007-biographies,ziser-reichart-2017-neural} to create domain-invariant representations of predictive features. Others learn low-dimensional transformation features of the data \cite{guo-etal-2009-domain,Glorot2011,Chen2012,yu-jiang-2016-learning,barnes-etal-2018-projecting}.   Finally, some approaches divide the feature space into general and domain-dependent features \cite{daume-iii-2007-frustratingly}. Multi-task learning can also improve cross-domain performance \cite{peng-dredze-2017-multi}.

Recently, \newcite{han-eisenstein-2019-unsupervised} proposed \textit{domain-adaptive fine-tuning}, where contextualised embeddings are first fine-tuned to both the source and target domains with a language modelling loss and subsequently fine-tuned to source domain labelled data. This approach outperforms several strong baselines trained on the target domain of the WNUT 2016 NER task \cite{strauss-etal-2016-results}.

\paragraph{Aggregation of annotations:} Approaches that aggregate annotations from multiples sources have largely concentrated on noisy data from crowd sourced annotations, with some annotators possibly being adversarial. The \textit{Bayesian Classifier Combination} approach of \newcite{pmlr-v22-kim12} combines multiple independent classifiers using a linear combination of predictions. \newcite{hovy-etal-2013-learning} learn a generative model able to aggregate crowd-sourced annotations and estimate the trustworthiness of annotators. \newcite{Rodrigues:2014:SLM:2843614.2843687} present an approach based on Conditional Random Fields (CRFs) whose model parameters are learned jointly using EM. \newcite{nguyen-etal-2017-aggregating} propose a Hidden Markov Model to aggregate crowd-sourced sequence annotations and find that explicitly modelling the annotator leads to improvements for POS-tagging and NER.  Finally, \newcite{simpson-gurevych-2019-bayesian} proposed a fully Bayesian approach to the problem of aggregating multiple sequential annotations, using variational EM to compute posterior distributions over the model parameters.

\paragraph{Weak supervision:} The aim of weakly supervised modelling is to reduce the need for hand-annotated data in supervised training.  A particular instance of weak supervision is \textit{distant supervision}, which relies on external resources such as knowledge bases to automatically label documents with entities that are known to belong to a particular category \cite{mintz-etal-2009-distant,ritter-etal-2013-modeling,shang-etal-2018-learning}. \newcite{Ratner:2017:SRT:3173074.3173077,Ratner2019} generalised this approach with the Snorkel framework which combines various supervision sources using a generative model to estimate the accuracy (and possible correlations) of each source. These aggregated supervision sources are then employed to train a discriminative model. Current frameworks are, however, not easily adaptable to sequence labelling tasks, as they typically require data points to be independent. One exception is the work of \newcite{wang-poon-2018-deep}, which relies on deep probabilistic logic to perform joint inference on the full dataset. Finally, \newcite{fries2017swellshark} presented a weak supervision approach to NER in the biomedical domain. However, unlike the model proposed in this paper, their approach relies on an ad-hoc mechanism for generating candidate spans to classify.

The approach most closely related to this paper is \newcite{safranchik:aaai20}, which describe a similar weak supervision framework for sequence labelling based on an extension of HMMs called linked hidden Markov models. The authors introduce a new type of noisy rules, called linking rules, to determine how sequence elements should be grouped into spans of same tag. The main differences between their approach and this paper are the linking rules, which are not employed here, and the choice of labelling functions, in particular the document-level relations detailed in Section \ref{sec:labelfunctions}. 

\paragraph{Ensemble learning:} The proposed approach is also loosely related to \textit{ensemble methods} such bagging, boosting and random forests \cite{doi:10.1002/widm.1249}.  These methods rely on multiple classifiers run simultaneously and whose outputs are combined at prediction time. In contrast, our approach (as in other weak supervision frameworks) only requires labelling functions to be aggregated once, as an intermediary step to create training data for the final model. This is a non-trivial difference as running all labelling functions at prediction time is computationally costly due to the need to run multiple neural models along with gazetteers extracted from large knowledge bases.

\section{Approach}

\begin{figure}[t]
    \centering
    \includegraphics[width=.47\textwidth]{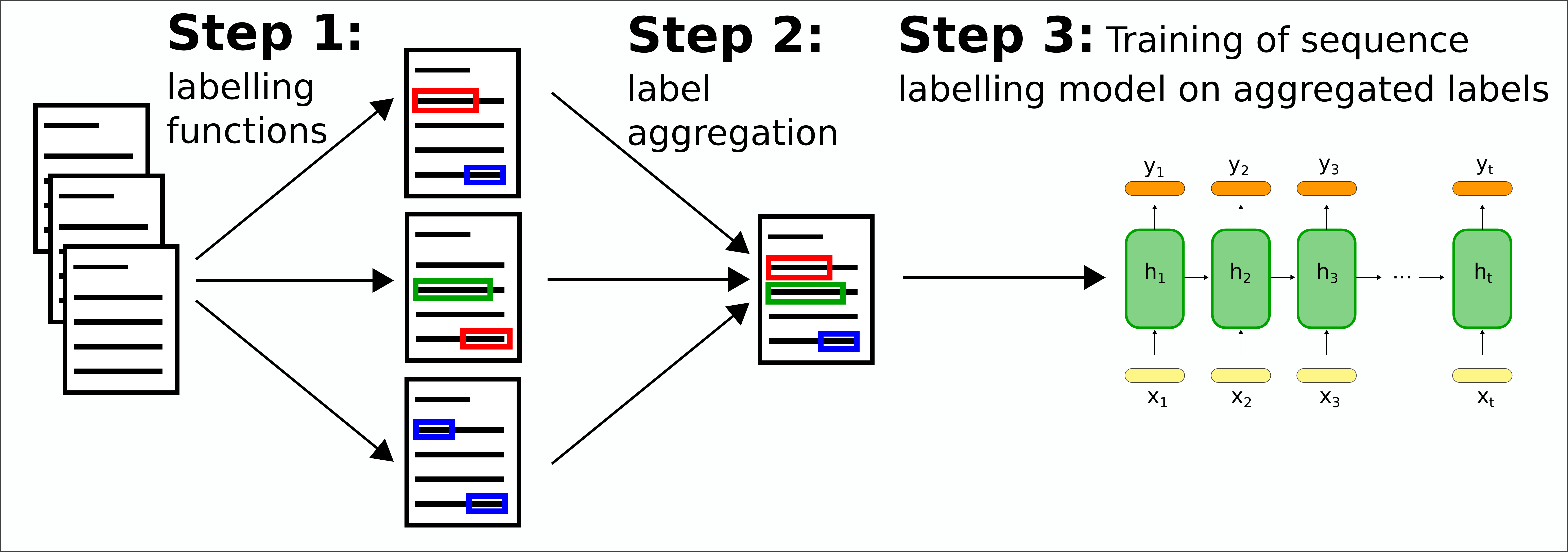}
    \caption{Illustration of the weak supervision approach.} \vspace{-12mm}
    \label{fig:approach}
\end{figure}

The proposed model collects weak supervision from multiple \textit{labelling functions}. Each labelling function takes a text document as input and outputs a series of spans associated with NER labels. These outputs are then aggregated using a hidden Markov model (HMM) with multiple emissions (one per labelling function) whose parameters are estimated in an unsupervised manner. Finally, the aggregated labels are employed to learn a sequence labelling model. Figure \ref{fig:approach} illustrates this process. The process is performed on documents from the target domain, e.g. a corpus of financial news.

Labelling functions are typically \textit{specialised} to detect only a subset of possible labels. For instance, a gazetteer based on Wikipedia will only detect mentions of persons, organisations and geographical locations and ignore entities such as dates or percents. This marks a departure from existing aggregation methods, which are originally designed for crowd-sourced data and where annotators are supposed to make use of the full label set. 
In addition, unlike previous weak supervision approaches, we allow labelling functions to produce \textit{probabilistic predictions} instead of deterministic values. The aggregation model described in Section \ref{sec:aggregation} directly captures these properties in the emission model associated with each labelling function. 

We first briefly describe the labelling functions integrated into the current system. We review in Section \ref{sec:aggregation} the aggregation model employed to combine the labelling predictions. The final labelling model is presented in Section \ref{sec:seqlabelling}.  The complete list of 52 labelling functions employed in the experiments is available in Appendix A.

\subsection{Labelling functions}
\label{sec:labelfunctions}

\paragraph{Out-of-domain NER models}
\label{sec:nermodels}

The first set of labelling functions are sequence labelling models trained in domains from which labelled data is available. In the experiments detailed in Section \ref{sec:expsetup}, we use four such models, respectively trained on Ontonotes \cite{ontonotes2011}, CoNLL 2003 \cite{tjong-kim-sang-de-meulder-2003-introduction}\footnote{The ConLL 2003 NER model is of course deactivated for the experimental evaluation on ConLL 2003.}, the Broad Twitter Corpus \cite{derczynski-etal-2016-broad} and a NER-annotated corpus of SEC filings \cite{salinas-alvarado-etal-2015-domain}. 

For the experiments in this paper, all aforementioned models rely on a transition-based NER model \cite{lample-etal-2016-neural} which extracts features with a stack of four convolutional layers with filter size of three and residual connections. The model uses attention features and a multi-layer perceptron to select the next transition. It is initialised with GloVe embeddings \cite{pennington-etal-2014-glove} and implemented in Spacy \cite{spacy2}. However, the proposed approach does not impose any constraints on the model architecture and alternative approaches based on e.g.~contextualised embeddings can also be employed. 

\paragraph{Gazetteers}

As in distant supervision approaches, we include a number of gazetteers from large knowledge bases to identify named entities. Concretely, we use resources from Wikipedia \cite{10.1007/978-3-319-73706-5_10}, Geonames \cite{wick2015geonames}, the Crunchbase Open Data Map, DBPedia \cite{journals/semweb/LehmannIJJKMHMK15} along with lists of countries, languages, nationalities and religious or political groups. 

To efficiently search for occurrences of these entities in large text collections, we first convert each knowledge base into a \textit{trie} data structure. Prefix search is then applied to extract matches (using both case-sensitive and case-insensitive mode, as they have distinct precision-recall trade-offs).

\paragraph{Heuristic functions}

We also include various heuristic functions, each specialised in the recognition of specific types of named entities. Several functions are dedicated to the recognition of proper names based on casing, part-of-speech tags or dependency relations. In addition, we integrate a variety of handcrafted functions relying on regular expressions to detect occurrences of various entities (see Appendix A for details). A probabilistic parser specialised in the recognition of dates, times, money amounts, percents, and cardinal/ordinal values \cite{braun-etal-2017-evaluating} is also incorporated. 

\paragraph{Document-level relations}

All labelling functions described above rely on local decisions on tokens or phrases. However, texts are not loose collections of words, but exhibit a high degree of internal coherence \cite{grosz-sidner-1986-attention,grosz-etal-1995-centering} which can be exploited to further improve the annotations.

We introduce one labelling function to capture \textit{label consistency} constraints in a document. As noted in \cite{krishnan-manning-2006-effective, Wang:2018:DNE:3229525.3213544}, named entities occurring multiple times through a document have a high probability of belonging to the same category. For instance, while \textit{Komatsu} may both refer to a Japanese town or a multinational corporation, a text including this mention will either be about the town or the company, but rarely both at the same time. 
To capture these non-local dependencies, we define the following label consistency model: given a text span $e$ occurring in a given document, we look for all spans $Z_e$ in the document that contain the same string as $e$. The (probabilistic) output of the labelling function then corresponds to the relative frequency of each label $l$ for that string in the document: 
\begin{equation}
P_{{\scriptsize \textsf{doc\_majority}(e)}}(l) = \frac{\sum_{z \in Z_e} P_{{\scriptsize \textsf{label}(z)}}(l)}{|Z_e|}
\end{equation}
The above formula depends on a distribution $P_{{\scriptsize \textsf{label}(z)}}$,  which can be defined on the basis of other labelling functions. Alternatively, a two-stage model similar to \cite{krishnan-manning-2006-effective} could be employed to first aggregate local labelling functions and subsequently apply document-level functions on aggregated predictions.

Another insight from \newcite{grosz-sidner-1986-attention} is the importance of the \textit{attentional structure}. When introduced for the first time, named entities are often referred to in an explicit and univocal manner, while subsequent mentions (once the entity is a part of the focus structure) frequently rely on shorter references.
The first mention of a person in a given text is for instance likely to include the person's full name, and is often shortened to the person's last name in subsequent mentions.
As in \newcite{ratinov-roth-2009-design}, we determine whether a proper name is a substring of another entity mentioned earlier in the text.  If so, the labelling function replicates the label distribution of the first entity.  

\subsection{Aggregation model}
\label{sec:aggregation}

The outputs of these labelling functions are then aggregated into a single layer of annotation through an \textit{aggregation model}. As we do not have access to labelled data for the target domain, this model is estimated in a fully unsupervised manner.

\paragraph{Model} We assume a list of $J$ labelling functions $\{\lambda_1,... \lambda_J\}$ and a list of $S$ mutually exclusive NER labels $\{l_1,...l_S\}$. The aggregation model is represented as an HMM, in which the states correspond to the true underlying labels. This model has multiple emissions (one per labelling function) assumed to be mutually independent conditional on the latent underlying label. 

\begin{figure}[h]
\centering
\tikzstyle{token0}=[shape=rectangle]
\tikzstyle{token}=[shape=rectangle, draw=black]
\tikzstyle{alpha}=[shape=circle, draw=black]
\tikzstyle{state}=[shape=circle, draw=black]
\tikzstyle{observation}=[shape=circle,draw=black, fill=lightgray]
\usetikzlibrary{bayesnet}

\vspace{3mm}
\begin{tikzpicture}[]
\node[token] (tim) at (0,3) {The} ;
\node[token] (ti) at (1.5,2.96) {plugged} ;
\node[token] (tip) at (3,3) {wells} ;
\node[token] (tipp) at (4.5,3) {have} ;
\node[token0] (tippp) at (6,3) {...} ;
\node[state] (sim) at (0,2) {$s_{i-1}$} ;
\node[state] (si) at (1.5,2) {$\ s_{i\ } \ $}
 edge [<-] (sim) ;
\node[state] (sip) at (3,2) {$s_{i+1}$}
 edge [<-] (si) ;
\node[state] (sipp) at (4.5,2) {$s_{i+2}$}
 edge [<-] (sip) ;
\node[token0] (sippp) at (6,2) {...}
 edge [<-] (sipp) ;
 \node[alpha] (alphai) at (1.5,0) {$\boldsymbol \alpha_{j}^{s_i}$} 
 edge [<-] (si) ;
 \node[observation] (pij) at (3.5,0) {$\boldsymbol P_{ij}$} 
 edge [<-] (alphai) ;
 \plate[inner sep=.25cm] {palphai} {(alphai)(pij)} {Labelling function $j \in \{1,... J\}$} ;
\end{tikzpicture}
\caption{Aggregation model using a hidden Markov model with multiple probabilistic emissions.}
    \label{fig:hmm} \vspace{-10mm}
\end{figure}
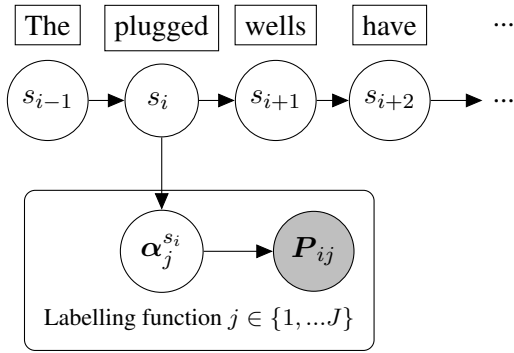

Formally, for each token $i \in \{1,...,n\}$ and labelling function $j$, we assume a Dirichlet distribution for the probability labels $\boldsymbol{P_{ij}}$. The parameters of this Dirichlet are separate vectors $\boldsymbol{\alpha_j^{s_i}}\in\mathcal{R}^S_{[0,1]}$, for each of the latent states $s_i \in \{1,...,S\}$. The latent states are assumed to have a Markovian dependence structure between the tokens $\{1,...,n\}$. This results in the HMM represented by a dependent mixtures of Dirichlet model:
\begin{align}
 &\boldsymbol{P_{ij}}|\boldsymbol{\alpha^{s_i}_j} \quad \overset{ind}{\sim}   \quad \text{Dirichlet}\left(\boldsymbol{\alpha_j^{s_i}}\right),\label{themodelbeg}
\\   &p(s_i|s_{i-1})\quad =\quad \text{logit}^{-1}\left(\omega^{(s_i,s_{i-1})}\right),\label{themodelmid} \\
&\text{logit}^{-1}\left(\omega^{(s_i,s_{i-1})}\right) =\tfrac{e^{\omega^{(s_i,s_{i-1})}}}{1+e^{\omega^{(s_i,s_{i-1})}}}.\label{themodeleqend}
\end{align}
Here, $\omega^{(s_i,s_{i-1})} \in \mathcal{R}$ are the parameters of the transition probability matrix controlling for a given state $s_{i-1}$ the probability of transition to state $s_{i}$.  Figure \ref{fig:hmm} illustrates the model structure. 

\paragraph{Parameter estimation} The learnable parameters of this HMM are (a) the transition matrix between states and (b) the $\boldsymbol{\alpha}$ vectors of the Dirichlet distribution associated with each labelling function. The transition matrix is of size $|S| \times |S|$, while we have $|S| \times |J|$ $\boldsymbol{\alpha}$ vectors, each of size $|S|$. The parameters are estimated with the Baum-Welch algorithm, which is a variant of EM algorithm that relies on the forward-backward algorithm to compute the statistics for the expectation step.   

To ensure faster convergence, we introduce a new constraint to the likelihood function: for each token position $i$, the corresponding latent label $s_i$ must have a non-zero probability in at least one labelling function (the likelihood of this label is otherwise set to zero for that position).  In other words, the aggregation model will only predict a particular label if this label is produced by least one labelling function. This simple constraint facilitates EM convergence as it restricts the state space to a few possible labels at every time-step. 

\paragraph{Prior distributions}

The HMM described above can be provided with informative priors. In particular, the initial distribution for the latent states can be defined as a Dirichlet based on counts $\delta$ for the most reliable labelling function\footnote{The most reliable labelling function was found in our experiments to be the NER model trained on Ontonotes 5.0.}:
\begin{align}
  &p(s_i) \overset{d}{=}   \text{Dirichlet}(\boldsymbol{\delta}).
\end{align}
The prior for each row $k$ of the transition probabilities matrix is also a Dirichlet based on the frequencies of transitions between the observed classes for the most reliable labelling function $\boldsymbol\kappa_k$:
\begin{align}
      &p(s_i|s_{i-1} = k) \overset{d}{=} \text{Dirichlet}({\boldsymbol\kappa_k}).
\end{align}

Finally, to facilitate convergence of the EM algorithm, informative starting values can be specified for the emission model of each labelling function. Assuming we can provide rough estimates of the recall $r_{jk}$ and precision $\rho_{jk}$ for the labelling function $j$ on label $k$, the initial values for the parameters of the emission model are expressed as: 
\begin{align}
    & \alpha_{jk}^{s_i}\propto
    \begin{cases}
r_{jk}, \text{ if } s_i=k,\\
\left(1-r_{s_ik}\right)  \left(1-\rho_{jk}\right) \delta_k,\text{ if } s_i \neq k.
\end{cases}\nonumber
\end{align}

The probability of observing a given label $k$ emitted by the labelling function $j$ is thus proportional to its recall if the true label is indeed $k$.  Otherwise (i.e.~if the labelling function made an error), the probability of emitting $k$ is inversely proportional to the precision of the labelling function $j$.

\paragraph{Decoding}

Once the parameters of the HMM model are estimated, the forward-backward algorithm can be employed to associate each token marginally with a posterior probability distribution over possible NER labels \cite{Rabiner:1990:THM:108235.108253}. 

\subsection{Sequence labelling model}
\label{sec:seqlabelling}

Once the labelling functions are aggregated on documents from the target domain, we can train a sequence labelling model on the unified annotations, without imposing any constraints on the type of model to use. To take advantage of the posterior marginal distribution ${\tilde p_{s}}$ over the latent labels, the optimisation should seek to minimise the expected loss with respect to ${\tilde p_{s}}$:
\begin{equation}
    \hat{\theta} = \argmin_{\theta} \sum_i^n \mathbb{E}_{y \sim {\tilde p_{s}}} \left[loss(h_{\theta}(x_i), y)\right]
\end{equation}
where $h_{\theta}(\cdot)$ is the output of the sequence labelling model. This is equivalent to minimising the cross-entropy error between the outputs of the neural model and the probabilistic labels produced by the aggregation model.

\section{Evaluation}
\label{sec:expsetup}

We evaluate the proposed approach on two English-language datasets, namely the CoNLL 2003 dataset and a collection of sentences from Reuters and Bloomberg news articles annotated with named entities by crowd-sourcing. We include a second dataset in order to evaluate the approach with a more fine-grained set of NER labels than the ones in CoNLL 2003.  As the objective of this paper is to compare approaches to unsupervised domain adaptation, we do not rely on any labelled data from these two target domains.

\subsection{ Data}
\label{sec:data}

\paragraph{CoNLL 2003} The CoNLL 2003 dataset \cite{tjong-kim-sang-de-meulder-2003-introduction} consists of 1163 documents, including a total of 35089 entities spread over 4 labels: \nerlabel{ORG}, \nerlabel{PER}, \nerlabel{LOC} and \nerlabel{MISC}.

\paragraph{Reuters \& Bloomberg} We additionally crowd annotate 1054 sentences from Reuters and Bloomberg news articles from \newcite{ding-etal-2014-using}. 
We instructed the annotators to tag sentences with the following 9 Ontonotes-inspired labels: \nerlabel{PERSON}, \nerlabel{NORP}, \nerlabel{ORG}, \nerlabel{LOC}, \nerlabel{PRODUCT}, \nerlabel{DATETIME}, \nerlabel{PERCENT}, \nerlabel{MONEY}, \nerlabel{QUANTITY}. Note that the \nerlabel{DATE} and \nerlabel{TIME} labels from Ontonotes are merged into \nerlabel{DATETIME}, and the \nerlabel{LOC} and \nerlabel{GPE} labels are similarly merged into \nerlabel{LOC}. Each sentence was annotated by at least two annotators, and a qualifying test with gold-annotated questions was conducted for quality control. Cohen's $\kappa$ for sentences with two annotators is 0.39, while Krippendorff's $\alpha$ for three annotators is 0.44. We had to remove  \nerlabel{QUANTITY} labels from the annotations as the crowd results for this particular label were highly inconsistent. 

\subsection{Baselines}
\label{sec:baselines}

\paragraph{Ontonotes-trained NER} The first baseline corresponds to a neural sequence labelling model trained on the Ontonotes 5.0 corpus. We use here the same model from Section \ref{sec:nermodels}, which is the single best-performing labelling function (that is, without aggregating multiple predictions). 

We also experimented with other neural architectures but these performed similar or worse than the transition-based model, presumably because they are more prone to overfitting on the source domain.

\paragraph{Majority voting (MV)}

The simplest method for aggregating outputs is majority voting, i.e.~outputting the most frequent label among the ones predicted by each labelling function. However, specialised labelling functions will output \nerlabel{O} for most tokens, which means that the majority label is typically \nerlabel{O}. 
To mitigate this problem, we first look at tokens that are marked with a non-\nerlabel{O} label by at least $T$ labelling functions (where $T$ is a hyper-parameter tuned experimentally), and then apply majority voting on this set of non-\nerlabel{O} labels. 

\paragraph{Snorkel model}

The Snorkel framework \cite{Ratner:2017:SRT:3173074.3173077} does not directly support sequence labelling tasks as data points are required to be independent. However, heuristics can be used to extract named-entity candidates and then apply labelling functions to infer their most likely labels \cite{fries2017swellshark}. For this baseline, we use the three functions \nerlabel{nnp\_detector}, \nerlabel{proper\_detector} and \nerlabel{compound\_detector} (see Appendix A) to generate candidate spans. We then create a matrix expressing the prediction of each labelling function for each span (including a specific "abstain" value to denote the absence of predictions) and run the matrix-completion-style approach of \newcite{Ratner2019} to aggregate the predictions. 

\paragraph{mSDA} is a strong domain adaptation baseline \cite{Chen2012} which augments the feature space of a model with intermediate representations learned using stacked denoising autoencoders. In our case, we learn the mSDA representations on the unlabeled source and target domain data. These 800 dimensional vectors are concatenated to 300 dimensional word embeddings and fed as input to a two-layer LSTM with a skip connection. Finally, we train the LSTM on the labeled source data and test on the target domain.

\paragraph{AdaptaBERT}

This baseline corresponds to a state-of-the-art unsupervised domain adaptation approach (AdaptaBERT) \cite{han-eisenstein-2019-unsupervised}. The approach first uses unlabeled data from both the source and target domains to domain-tune a pretrained BERT model. The model is finally task-tuned in a supervised fashion on the source domain labelled data (Ontonotes). At inference time, the model is able to make use of the pretraining and domain tuning to predict entities in the target domain. In our experiments, we use the cased-version of the base BERT model (trained on Wikipedia and news text) and perform three fine-tuning epochs for both domain-tuning and task-tuning. We additionally include an ensemble model, which averages the predictions of five BERT models fine-tuned with different random seeds.

\subsubsection*{Mixtures of multinomials}
Following the notation from Section~\ref{sec:aggregation}, we define $Y_{i,j,k} = \text{I}(P_{i,j,k}=\max_{k'\in\{1,...,S\}}P_{i,j,k'})$ to be the most probable label for word $i$ by source $j$. One can model $\boldsymbol{Y_{ij}}$ with a Multinomial probability distribution. The first four baselines (the fifth one assumes Markovian dependence between the latent states) listed below use the following independent, i.e. $p(s_i,s_{i-1}) = p(s_i)p(s_{i-1})$, mixtures of Multinomials model for $\boldsymbol{Y_{ij}}$:
\begin{align*}
 \boldsymbol{Y_{ij}}|\boldsymbol{p^{s_i}_j}& \quad \overset{ind}{\sim}   \quad \text{Multinomial}(\boldsymbol{p_j^{s_i}}),\\
\boldsymbol s_i&\quad \overset{ind}{\sim}  \quad\text{Multinomial}(\boldsymbol{\sigma}).
\end{align*}

\paragraph{Accuracy model (ACC)} 
\cite{Rodrigues:2014:SLM:2843614.2843687} assumes the following constraints on $\boldsymbol{p^{s_i}_j}$:
\begin{align*}
    & p_{jk}^{s_i} = \begin{cases}
\pi_j, \text{ if } s_i=k,\\
\tfrac{1-\pi_j}{J-1} s_i \neq k.
\end{cases}\nonumber
\end{align*}
Here, for each labelling function it is assumed to have the same accuracy $\pi_j$ for all of the tokens.
\paragraph{Confusion vector (CV)} \cite{nguyen2017aggregating} extends \textbf{ACC} by relying on separate success probabilities for each token label:
\begin{align*}
    & p_{jk}^{s_i} = \begin{cases}
\pi_{jk}, \text{ if } s_i=k,\\
\tfrac{1-\pi_{jk}}{J-1} s_i \neq k.
\end{cases}\nonumber
\end{align*}

\paragraph{Confusion matrix (CM)} \cite{Dawid:Skene:79} allows for distinct accuracies 
conditional on the latent states, which results in:
\begin{align}
    & p_{jk}^{s_i} = \pi_{jk}^{s_i}.\label{cmeq}
 \end{align}

\paragraph{Sequential Confusion Matrix (SEQ)} extends the \textit{CM} model of \citet{simpson-gurevych-2019-bayesian}, where an "auto-regressive" component is included in the observed part of the model. We assume dependence on a covariate indicating that the label has not changed for a given source, i.e.:
\begin{align*}
    & p_{jk}^{s_i} = \text{logit}^{-1}(\mu_{jk}^{s_i} +
    \text{I}(Y_{i-1,j,k}^{T}=Y_{i,j,k}^{T}){\beta}_{jk}^{s_i}).\nonumber
\end{align*}
\paragraph{Dependent confusion matrix (DCM)} combines the CM-distinct accuracies 
conditional on the latent states of \eqref{cmeq} and the Markovian dependence of \eqref{themodelmid}.

\subsection{Results}
\label{sec:results}

\begin{table*}[h!]
    \centering
    \begin{tabular}{p{7cm}rrrlrrr}

& \multicolumn{4}{c|}{Token-level} & \multicolumn{3}{c}{Entity-level} \\ 
 Model: &  P &  R &  $F_1$ & CEE &  P &  R &  $F_1$ \\
\midrule
Ontonotes-trained NER        &            0.719 &         0.706 &     0.712 &     2.671 &             0.694 &          0.620 &      0.654 \\[2mm]
Majority voting (MV)       &            \textbf{0.815} &         0.675 &     0.738 &     2.047 &             \textbf{0.751 }&          0.619 &      0.678 \\
Confusion Matrix (CM) &0.786&0.746&\textbf{0.766}&1.964&0.713&0.700&0.706\\
Sequential Confusion Matrix (SEQ) &0.736&0.716&0.726&2.254&0.642&0.668&0.654\\
Dependent Confusion Matrix (DCM) &0.785&0.744&0.764&1.983&0.710&0.698&0.704\\
Snorkel-aggregated labels      &            0.710 &         0.661 &     0.684 &     2.264 &             0.714 &          0.621 &      0.664 \\[2mm]
mSDA (OntoNotes) & 0.640 & 0.569 & 0.603 & 2.813 & 0.560 & 0.562 & 0.561 \\
AdaptaBERT (OntoNotes) & 0.693 & 0.733 & 0.712 & 2.280 & 0.652 & 0.736 & 0.691 \\
AdaptaBERT (Ensemble) & 0.704 & 0.754 & 0.729 & 2.103 & 0.684 & \textbf{0.743} & 0.712 \\[2mm]
HMM-aggregated labels (only NER models)     &            0.658 &         0.720 &     0.688 &     2.653 &             0.642 &          0.599 &      0.620 \\
HMM-aggregated labels (only gazetteers) &            0.759 &         0.394 &     0.518 &     3.678 &             0.687 &          0.367 &      0.478 \\
HMM-aggregated labels (only heuristics)    &            0.722 &         0.771 &     0.746 &     1.989 &             0.718 &          0.683 &      0.700 \\
HMM-aggregated labels (all but doc-level)    &            0.714 &         0.778 &     0.744 &     1.878 &             0.713 &          0.693 &      0.702 \\
HMM-aggregated labels (all functions)                 &            {0.719} &         \textbf{0.794} &     {0.754} &     \textbf{1.812} &             {0.721} &          0.713 &      \textbf{0.716} \\[2mm]
Neural net trained on HMM-agg. labels         &            0.712 &         0.790 &     0.748 &     2.282 &             0.715 &          0.707 &      0.710 \\[1mm]

\bottomrule
\end{tabular}
    \caption{Evaluation results on CoNLL 2003. MV=Majority Voting, P=Precision, R=Recall, CEE=Cross-entropy Error (lower is better). The results are micro-averaged on all labels (\nerlabel{PER}, \nerlabel{ORG}, \nerlabel{LOC} and \nerlabel{MISC}).}
    \label{fig:conll_results}
\end{table*}

\begin{table*}[h!]
    \centering
    \begin{tabular}{p{7cm}rrrlrrr}
 & \multicolumn{4}{c|}{Token-level} & \multicolumn{3}{c}{Entity-level} \\ 
Model: &  P &  R &  $F_1$ & CEE &  P &  R &  $F_1$ \\
\midrule
OntoNotes-trained NER &    0.793 &         0.791 &     0.792 &     2.648 &             0.694 &          0.635 &      0.664 \\[2mm]
Majority voting (MV)      &         \textbf{0.832} &         0.713 &     0.768 &     2.454 &             0.699 &          0.644 &       0.670 \\
Confusion Matrix (CM) &0.816&0.702&0.754&2.708&0.667&0.636&0.652\\
Sequential Confusion Matrix (SEQ) &0.741&0.630&0.682&3.261&0.535&0.547&0.540\\
Dependent Confusion Matrix (DCM) & 0.819&0.706&0.758&2.702&0.673&0.641&0.656\\
[2mm]
mSDA (OntoNotes) & 0.749 & 0.751 & 0.750 & 2.501 & 0.618 & 0.684 & 0.649\\
 AdaptaBERT (OntoNotes) & 0.799 & 0.801 & 0.800 & 2.351 & 0.668 & 0.734 & 0.699 \\
 AdaptaBERT (Ensemble) & {0.813} & 0.815 & 0.814 & 2.265 & 0.682 & \textbf{0.748} & 0.713\\[2mm]
HMM-aggregated labels (all functions)         &           0.804 &         0.823 &     0.814 &     \textbf{2.219} &    {0.749} &         0.697 &      {0.722} \\[2mm]
Neural net trained on HMM-agg. labels  &            0.805 &         \textbf{0.827} &    \textbf{0.816} &     2.448 &             {0.749} &          0.701 &      \textbf{0.724}     \\[1mm]
\bottomrule
\end{tabular}
    \caption{Evaluation results on 1094 crowd-annotated sentences from Reuters and Bloomberg news articles. The results are micro-averaged on 8 labels (\nerlabel{PERSON}, \nerlabel{NORP}, \nerlabel{ORG}, \nerlabel{LOC}, \nerlabel{PRODUCT}, \nerlabel{DATE}, \nerlabel{PERCENT}, and \nerlabel{MONEY}).}
    \label{fig:reuters_bloomberg}
\end{table*}

The evaluation results are shown in Tables \ref{fig:conll_results} and \ref{fig:reuters_bloomberg}, respectively for the CoNLL 2003 data and the sentences extracted from Reuters and Bloomberg.  The metrics are the (micro-averaged) precision, recall and $F_1$ scores at both the token-level and entity-level. In addition, we indicate the token-level cross-entropy error (in log-scale).   As the labelling functions are defined on a richer annotation scheme than the four labels of ConLL 2003, we map \nerlabel{GPE} to \nerlabel{LOC} and \nerlabel{EVENT}, \nerlabel{FAC}, \nerlabel{LANGUAGE}, \nerlabel{LAW}, \nerlabel{NORP}, \nerlabel{PRODUCT} and \nerlabel{WORK\_OF\_ART} to \nerlabel{MISC}. 

The results for the {\bf ACC} and {\bf CV} baselines are not included in the two tables as the parameter estimation did not converge (and thus did not provide reliable estimates of the parameters). 

Table \ref{fig:conll_results} further details the results obtained using only a subset of labelling functions. Of particular interest is the positive contribution of document-level functions, boosting the entity-level $F_1$ from 0.702 to 0.716. This highlights the importance of document-level relations in NER. 

The last line of the two tables reports the performance of the neural sequence labelling model (described in Section \ref{sec:seqlabelling}) trained on the basis of the aggregated labels. We observe that the performance of this neural model remains close to the performance of the HMM-aggregated labels. This result shows that the knowledge from the labelling functions can be injected into a standard neural model without substantial loss.

\subsection{Discussion}
\label{sec:analysis}

Although not shown in the results due to space constraints, we also analysed whether the informative priors described in Section \ref{sec:aggregation} influenced the performance of the aggregation model. We found informative and non-informative priors to yield similar performance for CoNLL 2003. However, the performance of non-informative priors was very poor on the Reuters and Bloomberg sentences ($F_1$ at 0.12), thereby demonstrating the usefulness of informative priors for small datasets. 

\begin{figure*}[t]\centering
    \includegraphics[width=1.0\textwidth]{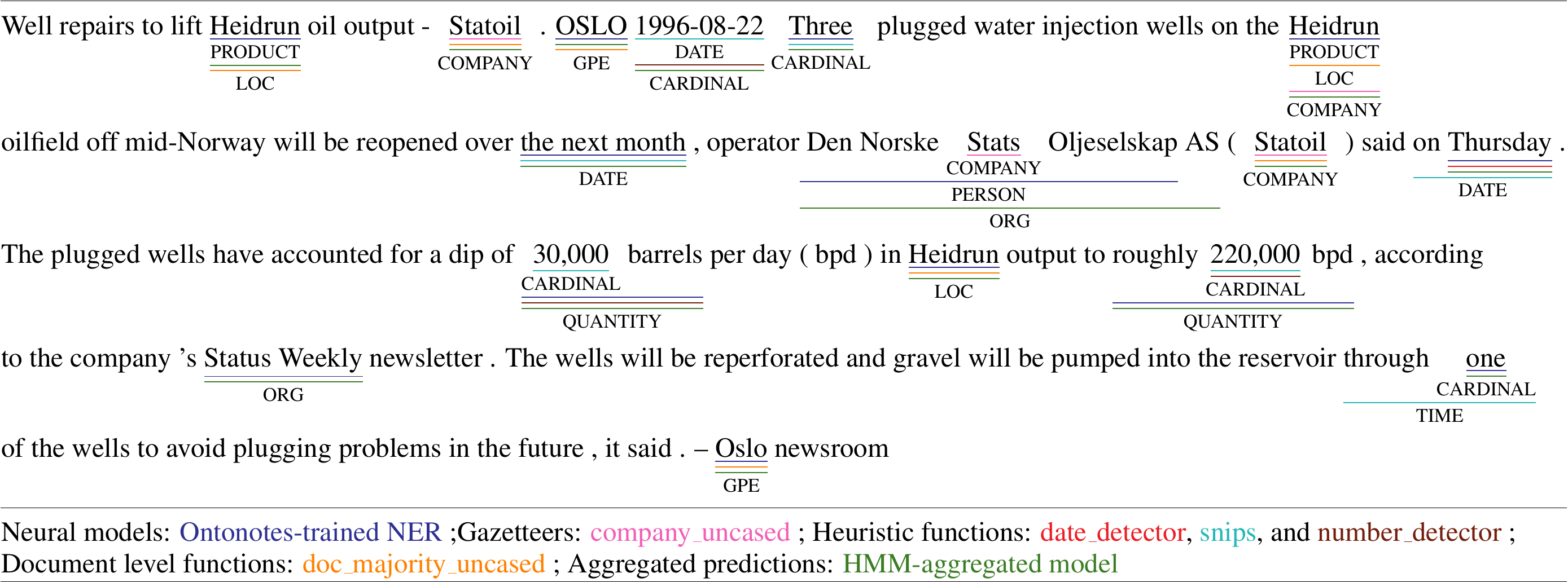}
    \caption{Extended example showing the outputs of 6 labelling functions, along with the HMM-aggregated model.}
    \label{fig:example}
\end{figure*}

We provide in Figure \ref{fig:example} an example with a few selected labelling functions. In particular, we can observe that the Ontonotes-trained NER model mistakenly labels "Heidrun" as a product. This erroneous label, however, is counter-balanced by other labelling functions, notably a document-level function looking at the global label frequency of this string through the document. We do, however, notice a few remaining errors, e.g. the labelling of "Status Weekly" as an organisation.

\begin{figure*}[t]
    \centering
    \includegraphics[width=.8\textwidth]{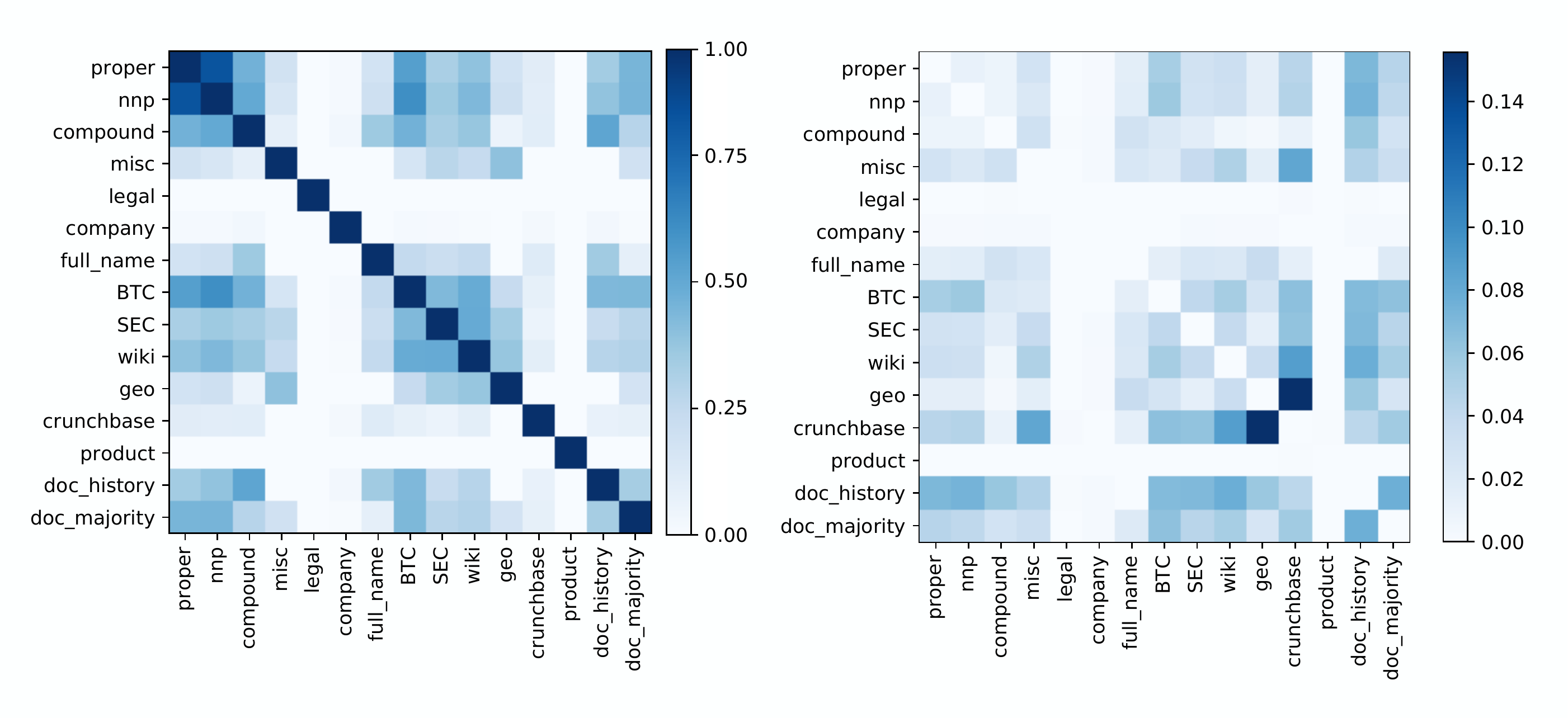} \vspace{-2mm}
    \caption{Pairwise agreement (left) and disagreement (right) between the labelling functions on the CoNLL 2003 data with labels \nerlabel{PER}, \nerlabel{ORG}, \nerlabel{LOC}, \nerlabel{MISC}, normalized by total number of labelled examples.}
    \label{fig:pairwiseagreement}
\end{figure*}

Figure \ref{fig:pairwiseagreement} illustrates the pairwise agreement and disagreement between labelling functions on the CoNLL 2003 dataset. If both labelling functions make the same prediction on a given token, we count this as an agreement, whereas conflicting predictions (ignoring \nerlabel{O} labels), are seen as disagreement. Large differences may exist between these functions for specific labels, especially \nerlabel{MISC}. The functions with the highest overlap are those making predictions on all labels, while labelling functions specialised to few labels (such as \nerlabel{legal\_detector}) often have less overlap. We also observe that the two gazetteers from Crunchbase and Geonames disagree in about 15\% of cases, presumably due to company names that are also geographical locations, as in the earlier Komatsu example. 

In terms of computational efficiency, the estimation of HMM parameters is relatively fast, requiring less than 30 mins on the entire CoNLL 2003 data.  Once the aggregation model is estimated, it can be directly applied to new texts with a single forward-backward pass, and can therefore scale to datasets with hundreds of thousands of documents. This runtime performance is an important advantage compared to approaches such as AdaptaBERT \cite{han-eisenstein-2019-unsupervised} which are relatively slow at inference time. The proposed approach can also be ported to other languages than English, although heuristic functions and gazetteers will need to be adapted to the target language. 

\section{Conclusion}
\label{sec:summary}

This paper presented a weak supervision model for sequence labelling tasks such as Named Entity Recognition. To leverage all possible knowledge sources available for the task, the approach uses a broad spectrum of labelling functions, including data-driven NER models, gazetteers, heuristic functions, and document-level relations between entities. Labelling functions may be specialised to recognise specific labels while ignoring others. Furthermore, unlike previous weak supervision approaches, labelling functions may produce probabilistic predictions. The outputs of these labelling functions are then merged together using a hidden Markov model whose parameters are estimated with the Baum-Welch algorithm. A neural sequence labelling model can finally be learned on the basis of these unified predictions.

Evaluation results on two datasets (CoNLL 2003 and news articles from Reuters and Bloomberg) show that the method can boost NER performance by about 7 percentage points on entity-level $F_1$. In particular, the proposed model outperforms the unsupervised domain adaptation approach through contextualised embeddings of \newcite{han-eisenstein-2019-unsupervised}. Of specific linguistic interest is the contribution of document-level labelling functions, which take advantage of the internal coherence and narrative structure of the texts.

Future work will investigate how to take into account potential correlations between labelling functions in the aggregation model, as done in e.g.~\cite{Bach:2017:LSG:3305381.3305410}. Furthermore, some of the labelling functions can be rather noisy and model selection of the optimal subset of the labelling functions might well improve the performance of our model. Model selection approaches that can be adapted are discussed in \citet{adams2019survey,hubin2019adaptive}.  We also wish to evaluate the approach on other types of sequence labelling tasks beyond Named Entity Recognition. 

\section*{Acknowledgements}

The research presented in this paper was conducted as part of the innovation project "FinAI: Artificial Intelligence tool to monitor global financial markets" in collaboration with Exabel AS\footnote{\url{www.exabel.com}}. This collaboration is supported through the funding programme for "User-driven Research based Innovation" of the Research Council of Norway. 

Additionally, this work is supported by the SANT project (Sentiment Analysis for Norwegian Text), funded by the Research Council of Norway (grant number 270908)

\bibliographystyle{acl_natbib}
\bibliography{anthology,acl2020}

\newpage
\appendix

\section{Labelling functions}
\label{sec:functionlist}

\noindent\begin{minipage}{\textwidth}
\begin{small}
\begin{tabular}{p{13mm}lp{99mm}} \\
\textbf{Group} & \textbf{Function name} & \textbf{Description} \\  \hline \\[-2mm]
\multirow{8}{15mm}{Neural NER models} & \labelfunsmall{BTC} &  Model trained on the Broad Twitter Corpus \\
& \labelfunsmall{BTC+c}& Model trained on the Broad Twitter Corpus + postprocessing  \\
& \labelfunsmall{SEC}&  Model trained on SEC-filings \\
& \labelfunsmall{SEC+c} & Model trained on SEC-filings + postprocessing \\
&\labelfunsmall{conll2003} &  Model trained on CoNLL 2003 \\
& \labelfunsmall{conll2003+c}  &  Model trained on CoNLL 2003 + postprocessing \\
&\labelfunsmall{core\_web\_md} &  Model trained on Ontonotes 5.0  \\
& \labelfunsmall{core\_web\_md+c} & Model trained on Ontonotes 5.0 + postprocessing \\[1mm] \hdashline \\[-2mm]
\multirow{24}{15mm}{Gazetteers} & \labelfunsmall{wiki\_cased}  & Gazetteer (case-sensitive) using Wikipedia entries \\
& \labelfunsmall{multitoken\_wiki\_cased} & Same as above, but restricted to multitoken entities \\
& \labelfunsmall{wiki\_uncased} & Gazetteer (case-insensitive) using Wikipedia entries \\
& \labelfunsmall{multitoken\_wiki\_uncased} &  Same as above, but restricted to multitoken entities\\ 
& \labelfunsmall{wiki\_small\_cased} &  Gazetteer (case-sensitive) using Wikipedia entries with non-empty description\\
& \labelfunsmall{multitoken\_wiki\_small\_cased} & Same as above, but restricted to multitoken entities \\
& \labelfunsmall{wiki\_small\_uncased} & Gazetteer (case-insensitive) using Wikipedia entries with non-empty description \\
& \labelfunsmall{multitoken\_wiki\_small\_uncased} & Same as above, but restricted to multitoken entities \\
& \labelfunsmall{company\_cased}& Gazetteer (case-sensitive) using a large list of company names  \\
&  \labelfunsmall{multitoken\_company\_cased} & Same as above, but restricted to multitoken entities \\
& \labelfunsmall{company\_uncased}&  Gazetteer from a large list of company names (case-insensitive) \\
& \labelfunsmall{multitoken\_company\_uncased} & Same as above, but restricted to multitoken entities \\
& \labelfunsmall{crunchbase\_cased} & Gazetteer (case-sensitive) using the Crunchbase Open Data Map \\
&\labelfunsmall{multitoken\_crunchbase\_cased} & Same as above, but restricted to multitoken entities \\
& \labelfunsmall{crunchbase\_uncased} & Gazetteer (case-insensitive) using the Crunchbase Open Data Map \\
& \labelfunsmall{multitoken\_crunchbase\_uncased} & Same as above, but restricted to multitoken entities \\
& \labelfunsmall{geo\_cased} & Gazetteer (case-sensitive) using the Geonames database \\
& \labelfunsmall{multitoken\_geo\_cased} & Same as above, but restricted to multitoken entities \\
& \labelfunsmall{geo\_uncased}  &  Gazetteer (case-insensitive) using the Geonames database \\
& \labelfunsmall{multitoken\_geo\_uncased} & Same as above, but restricted to multitoken entities \\
& \labelfunsmall{product\_cased}  & Gazetteer (case-sensitive) using products extracted from DBPedia \\
&\labelfunsmall{multitoken\_product\_cased} & Same as above, but restricted to multitoken entities \\
&\labelfunsmall{product\_uncased} &  Gazetteer (case-insensitive) using products extracted from DBPedia \\
&\labelfunsmall{multitoken\_product\_uncased} & Same as above, but restricted to multitoken entities  \\[1mm] \hdashline \\[-2mm]
 \multirow{17}{15mm}{Heuristic functions} & \labelfunsmall{date\_detector} & Detection of entities of type \nerlabelto{DATE} \\
&\labelfunsmall{time\_detector}  &  Detection of entities of type \nerlabelto{TIME}  \\
& \labelfunsmall{money\_detector} & Detection of entities of type \nerlabelto{MONEY} \\
& \labelfunsmall{number\_detector} & Detection of entities \nerlabelto{CARDINAL}, \nerlabelto{ORDINAL}, \nerlabelto{PERCENT} and \nerlabelto{QUANTITY} \\
&\labelfunsmall{legal\_detector} & Detection of entities of type \nerlabelto{LAW} \\
& \labelfunsmall{misc\_detector} &  Detection of entities of type \nerlabelto{NORP}, \nerlabelto{LANGUAGE}, \nerlabelto{FAC} or \nerlabelto{EVENT} \\
& \labelfunsmall{full\_name\_detector} & Heuristic function to detect full person names \\
& \labelfunsmall{company\_type\_detector}& Detection of companies with a legal type suffix \\
& \labelfunsmall{nnp\_detector} & Detection of sequences of tokens with \nerlabelto{NNP} as POS-tag\\
& \labelfunsmall{infrequent\_nnp\_detector} & Detection of sequences of tokens with \nerlabelto{NNP} as POS-tag \\ & & + including at least one infrequent token (rank $>$ 15000 in vocabulary)   \\
&\labelfunsmall{proper\_detector}  & Detection of proper names based on casing  \\
& \labelfunsmall{infrequent\_proper\_detector} &  Detection of proper names based on casing \\ & & +  including at least one infrequent token \\
& \labelfunsmall{proper2\_detector}  & Detection of proper names based on casing  \\
& \labelfunsmall{infrequent\_proper2\_detector} & Detection of proper names based on casing \\ & & +  including at least one infrequent token \\
& \labelfunsmall{compound\_detector} & Detection of proper noun phrases with compound dependency relations \\
&\labelfunsmall{infrequent\_compound\_detector} & Detection of proper noun phrases with compound dependency relations \\ & & +  including at least one infrequent token  \\
&\labelfunsmall{snips}  & Probabilistic parser specialised in the recognition of dates, times, money amounts, percents, and cardinal/ordinal values \\[1mm] \hdashline \\[-2mm]
\multirow{3}{15mm}{Document-level functions} & \labelfunsmall{doc\_history}  & Entity classification based on already introduced entities in the document \\
&\labelfunsmall{doc\_majority\_cased} &  Entity classification based on majority labels in document (case-sensitive) \\
& \labelfunsmall{doc\_majority\_uncased} &  Entity classification based on majority labels in document (case-insensitive) \\[3mm]
\end{tabular}
\captionof{table}{Full list of labelling functions employed in the experiments.  The neural NER models are provided in two versions: one that directly outputs the raw model predictions, and one that runs a shallow postprocessing step on the model predictions to correct known recognition errors (for instance, ensuring that a numeric amount that is either preceded or followed by a currency symbol is always classified as an entity of type \nerlabel{MONEY}). }

\end{small}
\end{minipage}

\newpage
$\phantom{x}$
\newpage

\section{Label matching problem}
The baseline models relying on mixtures of multinomials have to address the so-called \textit{label matching problem}, which needs some extra care. 

The following approach was employed in the experiments from Section 4: 
\begin{itemize}
    \item First, we put strong initial values to the probabilities $\boldsymbol{\sigma}$ of individual classes based on the frequency of appearance of these classes in the most reliable labelling function. This is expected to increase the probability of EM exploring the mode around the initialised values. 
    \item Second, we perform post-processing and set the labels to the states corresponding to the labels  with the highest pairwise correlations to the latent labels from one of the three options: 
    \begin{enumerate}
         \item the most reliable labelling function (Ontonotes-trained NER);
         \item the majority voting labelling function;
         \item the suggested Dirichlet dependent mixture model.
     \end{enumerate}
    Additionally, if this highest correlation is below the threshold of $0.1$ the \nerlabel{O} label is assigned to the corresponding state. We empirically observed that the label matching technique that performed best was to map the states to the labels produced by the majority voter (based on the pairwise correlations). 
\end{itemize}

\section{Detailed results}
\label{sec:detailed}

 In Table \ref{table:detailed}, we provide the detailed results distributed by NER label for the CoNLL data 2003 which were presented in micro-averaged form in Table 1 of the main paper.

\begin{table*}
\begin{tabular}{lllrrlrrr}
\toprule
Label & Proportion & Model & \multicolumn{3}{c|}{Token-level} & \multicolumn{3}{c}{Entity-level} \\ 
& & &  P &  R &  $F_1$ &  P &  R &  $F_1$ \\
\midrule
\nerlabel{LOC}  &  30.3 \%  & Ontonotes-trained NER  &            0.767 &         0.812 &     0.788 &             0.764 &          0.800 & 0.782 \\
         &        & Majority voting (MV) &            0.740 &         0.839 &     0.786 &             0.739 &          0.828 &      0.780 \\
&&Confusion Matrix&0.721&0.895&0.798&0.714&0.890&0.792\\
&&Sequential Confusion Matrix&0.681&0.856&0.758&0.664&0.848&0.744\\
&&Dependent Confusion Matrix&0.718&0.890&0.794&0.710&0.886&0.788\\
         &        & Snorkel-aggregated labels &            0.634 &         0.855 &     0.728 &             0.676 &          0.747 &      0.710 \\
                  &        & HMM (only NER models) &            0.601 &         0.825 &     0.696 &             0.650 &          0.733 &      0.690 \\
               &        & HMM (only gazetteers) &            0.707 &         0.632 &     0.668 &             0.694 &          0.630 &      0.660 \\
           &        & HMM (heuristics) &            0.715 &         0.870 &     0.784 &             0.745 &          0.832 &      0.786 \\
         & & HMM (all but doc-level) &            0.701 &         0.862 &     0.774 &             0.724 &          0.838 &      0.776 \\
         && HMM (all functions) &            0.726 &         0.859 &     0.786    &             0.738 &          0.839 &      0.786 \\
         &        & NN trained on HMM &            0.736 &         0.851 &     0.790 &             0.734 &          0.850 &      0.788 \\[2mm]
\nerlabel{PER} & 28.7 \% & Ontonotes-trained NER  &            0.850 &         0.833 &     0.842 &             0.787 &          0.741 &      0.764 \\
        &        & Majority voting (MV) &            0.915 &         0.871 &     0.892 &             0.831 &          0.775 &      0.802 \\
&&Confusion Matrix&0.891&0.921&0.906&0.806&0.834&0.820\\
&&Sequential Confusion Matrix&0.849&0.879&0.864&0.730&0.789&0.758\\
&&Dependent Confusion Matrix&0.892&0.920&0.906&0.806&0.834&0.820\\
         &        & Snorkel-aggregated labels &            0.816 &         0.903 &     0.858 &             0.769 &          0.717 &      0.742 \\
          &        & HMM (only NER models) &            0.837 &         0.860 &     0.848 &             0.770 &          0.744 &      0.756 \\
           &        & HMM (only gazetteers) &            0.917 &         0.452 &     0.606 &             0.835 &          0.391 &      0.532 \\
              &        & HMM (heuristics) &            0.836 &         0.933 &     0.882 &             0.791 &          0.799 &      0.794 \\
         && HMM (all but doc-level) &            0.859 &         0.917 &     0.888 &             0.814 &          0.782 &      0.798 \\
         && HMM (all functions) &            0.857 &         0.947 &     0.900      &             0.820 &          0.826 &      0.822 \\
         &        & NN trained on HMM &            0.856 &         0.946 &     0.898 &             0.814 &          0.824 &      0.818 \\[2mm]
\nerlabel{ORG} & 26.6 \%   & Ontonotes-trained NER  &            0.536 &         0.517 &     0.526 &             0.437 &          0.306 &      0.360 \\
         &        & Majority voting (MV) &            0.725 &         0.512 &     0.600 &             0.610 &          0.434 &      0.508 \\
&&Confusion Matrix&0.698&0.613&0.652&0.571&0.537&0.554\\         
&&Sequential Confusion Matrix&0.632&0.590&0.610&0.485&0.515&0.500\\
&&Dependent Confusion Matrix&0.696&0.613&0.652&0.567&0.536&0.552\\
         &        & Snorkel-aggregated labels &            0.512 &         0.639 &     0.568 &             0.519 &          0.496 &      0.508 \\
         &        & HMM (only NER models) &            0.516 &         0.549 &     0.532 &             0.425 &          0.333 &      0.374 \\
          &        & HMM (only gazetteers) &            0.648 &         0.304 &     0.414 &             0.512 &          0.235 &      0.322 \\
             &        & HMM (heuristics) &            0.566 &         0.625 &     0.594 &             0.549 &          0.501 &      0.524 \\
         && HMM (all but doc-level) &            0.565 &         0.631 &     0.596 &             0.551 &          0.494 &      0.520 \\
         && HMM (all functions) &            0.542 &         0.665 &     0.598    &             0.545 &          0.527 &      0.536 \\
         &        & NN trained on HMM &            0.539 &         0.665 &     0.596 &             0.537 &          0.519 &      0.528 \\[2mm]
\nerlabel{MISC}& 14.4 \% & Ontonotes-trained NER  &            0.676 &         0.599 &     0.636 &             0.702 &          0.583 &      0.636 \\
        &        & Majority voting (MV) &            0.861 &         0.187 &     0.308 &             0.809 &          0.193 &      0.312 \\
&&Confusion Matrix&0.895&0.319&0.470&0.850&0.332&0.478\\         
&&Sequential Confusion Matrix&0.850&0.320&0.464&0.791&0.333&0.468\\
&&Dependent Confusion Matrix&0.893&0.318&0.468&0.844&0.330&0.474\\

         &        & Snorkel-aggregated labels &            0.852 &         0.398 &     0.542 &             0.863 &          0.400 &      0.546 \\
       &        & HMM (only NER models) &            0.667 &         0.544 &     0.600 &             0.708 &          0.518 &      0.598 \\
          &        & HMM (only gazetteers) &            0.745 &         0.011 &     0.022 &             0.594 &          0.008 &      0.016 \\
        &        & HMM (heuristics) &            0.842 &         0.499 &     0.626 &             0.850 &          0.478 &      0.612 \\
         & & HMM (all but doc-level) &            0.714 &         0.596 &     0.650 &             0.781 &          0.575 &      0.662 \\
         && HMM (all functions) &            0.814 &         0.571 &     0.672     &             0.830 &          0.565 &      0.672 \\
         &        & NN trained on HMM &            0.852 &         0.577 &     0.688 &             0.866 &          0.583 &      0.696 \\
\bottomrule
\end{tabular}
\caption{Detailed evaluation results on the CoNLL2003 dataset, depending on NER labels.}
\label{table:detailed}
\end{table*}

\end{document}